\documentclass[sigconf]{acmart}

\usepackage[most]{tcolorbox}

\usepackage[normalem]{ulem}
\usepackage{tabularray}
\usepackage{makecell} 
\usepackage[normalem]{ulem}
\useunder{\uline}{\ul}{}
\usepackage{algorithm}
\usepackage{algorithmic}
\usepackage{multirow}
\usepackage{amsmath} 
\usepackage{enumitem}
\usepackage{balance}
\usepackage{caption} 

\usepackage{array} 
\usepackage{etoolbox} 
\usepackage{graphicx}
\usepackage{booktabs}

\usepackage{hyperref}
\usepackage{url}
\usepackage{amsmath} 
\usepackage{subcaption}
\usepackage{tcolorbox}
\usepackage{fancyvrb}
\usepackage{enumitem}
\usepackage{booktabs}       
\usepackage{amsfonts}       
\usepackage{nicefrac}       
\usepackage{microtype}      
\usepackage{xcolor}         
\usepackage{graphicx}
\usepackage{tcolorbox}
\usepackage{caption}

\copyrightyear{2026}
\acmYear{2026}
\setcopyright{cc}
\setcctype{by}
\acmConference[SIGIR '26]{Proceedings of the 49th International ACM SIGIR Conference on Research and Development in Information Retrieval}{July 20--24, 2026}{Melbourne, VIC, Australia}
\acmBooktitle{Proceedings of the 49th International ACM SIGIR Conference on Research and Development in Information Retrieval (SIGIR '26), July 20--24, 2026, Melbourne, VIC, Australia}
\acmDOI{10.1145/3805712.3809928}
\acmISBN{979-8-4007-2599-9/2026/07}

\begin{document}
\title{Enhancing Judgment Document Generation via Agentic Legal Information Collection and Rubric-Guided Optimization}

\author{Weihang Su}
\email{swh22@mails.tsinghua.edu.cn}
\affiliation{%
    \institution{Tsinghua University}
    \city{Beijing}
    \country{China}
}
\affiliation{%
    \institution{Quan Cheng Laboratory}
    \city{Shandong}
    \country{China}
}

\author{Xuanyi Chen}
\affiliation{%
    \institution{Tsinghua University}
    \city{Beijing}
    \country{China}
}

\author{Yueyue Wu}
\email{wuyueyue1600@gmail.com}
\affiliation{%
    \institution{Quan Cheng Laboratory}
    \city{Shandong}
    \country{China}
}
\affiliation{%
    \institution{Tsinghua University}
    \city{Beijing}
    \country{China}
}
\authornote{Corresponding Author}

\author{Qingyao Ai}
\email{aiqy@tsinghua.edu.cn}
\affiliation{%
    \institution{Quan Cheng Laboratory}
    \city{Shandong}
    \country{China}
}
\affiliation{%
    \institution{Tsinghua University}
    \city{Beijing}
    \country{China}
}
\authornotemark[1]

\author{Yiqun Liu}
\affiliation{%
    \institution{Tsinghua University}
    \city{Beijing}
    \country{China}
}

\begin{abstract}
Automating the drafting of judgment documents is pivotal to judicial efficiency, yet it remains challenging due to the dual requirements of comprehensive retrieval of legal information and rigorous logical reasoning. Existing approaches, typically relying on standard Retrieval-Augmented Generation and Supervised Fine-Tuning, often suffer from insufficient evidence recall, hallucinated statutory references, and logically flawed legal reasoning. To bridge this gap, we propose Judge-R1, a unified framework designed to enhance LLM-based judgment document generation by jointly improving legal information collection and judgment document generation. First, we introduce Agentic Legal Information Collection, which employs a dynamic planning agent to retrieve precise statutes and precedents from multiple sources. Second, we implement Rubric-Guided Optimization, a reinforcement learning phase utilizing Group Relative Policy Optimization (GRPO) with a comprehensive legal reward function to enforce adherence to judicial standards and reasoning logic. Extensive experiments on the JuDGE benchmark demonstrate that Judge-R1 significantly outperforms state-of-the-art baselines in both legal accuracy and generation quality\footnote{All the codes and datasets are available at: \url{github.com/oneal2000/Judge-R1}}.
\end{abstract}



\maketitle

\section{Introduction}
\label{sec:1}

Large Language Models (LLMs) have recently shown strong potential in legal applications, including contract analysis~\cite{hu2026evaluation}, case law research~\cite{ma2023caseencoder,su2023caseformer}, and legal document drafting~\cite{zhang2025chinese, su2025judge}. 
Among these applications, judgment document generation is a particularly important task at the intersection of legal information retrieval and text generation. 
To draft a judgment document, a system must identify case-relevant legal information, such as statutes and precedents, and integrate it with case facts through rigorous legal reasoning.
Unlike ordinary legal summaries, judgment documents are authoritative texts that record procedural considerations, reasoning processes, and final rulings, serving as a formal basis for judicial decisions~\cite{li2023sailer}. 
In practice, this process is labor-intensive for judges, who must synthesize diverse legal materials into a coherent and legally sound document. 
Automating this process, therefore, holds significant potential to reduce judicial workload and improve procedural efficiency.

However, despite the promise of generative AI and LLMs, high-quality judgment document generation remains an open challenge due to two fundamental bottlenecks: reliable information collection and legally grounded generation. 
First, the information-collection stage serves as the foundation of legal reasoning. 
To draft a valid judgment, a system must accurately retrieve the statutes and precedents that provide the legal authorities for constructing the major premise of legal reasoning.
This is non-trivial because legal queries are often verbose and abstract (derived from raw case facts), and the retrieval space is riddled with noise (e.g., distinguishing between semantically similar but legally distinct statutes). 
Failure to retrieve a key statute or relevant precedent inevitably results in a legally flawed output, regardless of the model's capabilities. 
Second, the document generation stage demands more than linguistic fluency. 
It requires the systematic integration of the retrieved legal knowledge with the factual minor premise to deduce a correct conclusion. The generated text must strictly adhere to judicial standards regarding sentencing logic, structural integrity, and professional tone.

Existing approaches often fall short in addressing these dual challenges. 
For information collection, standard Retrieval-Augmented Generation (RAG) methods typically rely on semantic matching, which lacks the planning capability to decompose complex case facts into precise legal queries, often resulting in insufficient recall or irrelevant noise. 
In the generation process, the prevailing paradigm largely relies on Supervised Fine-Tuning (SFT)~\cite{su2025judge}, which aims to help LLMs better imitate the structure, style, and reasoning patterns of legal texts. 
While SFT enables models to mimic the stylistic surface of legal texts, it fundamentally relies on imitation learning. 
It does not explicitly penalize logical fallacies or incentivize the model to revise its reasoning toward a more faithful and legally grounded judgment. 
As a result, models may produce judgments that are stylistically plausible but legally unreliable, featuring hallucinated citations, unsupported reasoning chains, or conclusions that are not grounded in legal authorities.

To this end, we introduce Judge-R1, a novel framework that bridges the gap between complex legal retrieval and rigorous judgment generation. 
Judge-R1 addresses these limitations through a synergistic approach. 
For information collection, we propose an Agentic Legal Information Collection mechanism. 
Unlike static retrievers, our method employs a Multi-source Retrieval-Augmented Generation (MRAG) system driven by an intelligent agent. 
This agent dynamically plans search queries to cover diverse legal points (e.g., constitutive elements, sentencing factors) and employs a multi-stage filter to construct a high-quality evidence set comprising both statutes and precedents. 
For Judgment Document Generation, we move beyond simple imitation by introducing a Reinforcement Learning (RL) alignment phase specifically designed for the legal domain. 
After a warm-up stage using SFT, we implement Group Relative Policy Optimization (GRPO) to further refine the model~\cite{guo2025deepseek}. 
Crucially, we design a comprehensive legal rubric as the reward function, which evaluates the output across multiple dimensions: legal accuracy (e.g., correct statutes and sentencing), textual quality, and reasoning depth. 
This allows the model to self-optimize its generation policy, learning to produce judgments that are not only linguistically coherent but also legally robust.

Extensive experiments on the JuDGE benchmark demonstrate that Judge-R1 outperforms strong baselines. For Information Collection, our agentic framework provides superior retrieval performance compared with standard RAG systems. For Judgment Document Generation, our RL-aligned approach yields judgments with significantly higher legal accuracy and logical consistency than existing methods.

In summary, our contributions are:
\begin{itemize}[leftmargin=*]
    \item We propose {Judge-R1}, a unified framework that integrates agentic information collection with rubric-driven reinforcement learning for judgment document generation.
    \item Empirical results show that Judge-R1 achieves state-of-the-art performance, offering a principled path for improving the reliability of legal text generation.
\end{itemize}

\begin{figure*}[t]
\centering
    \includegraphics[width=0.99\textwidth]{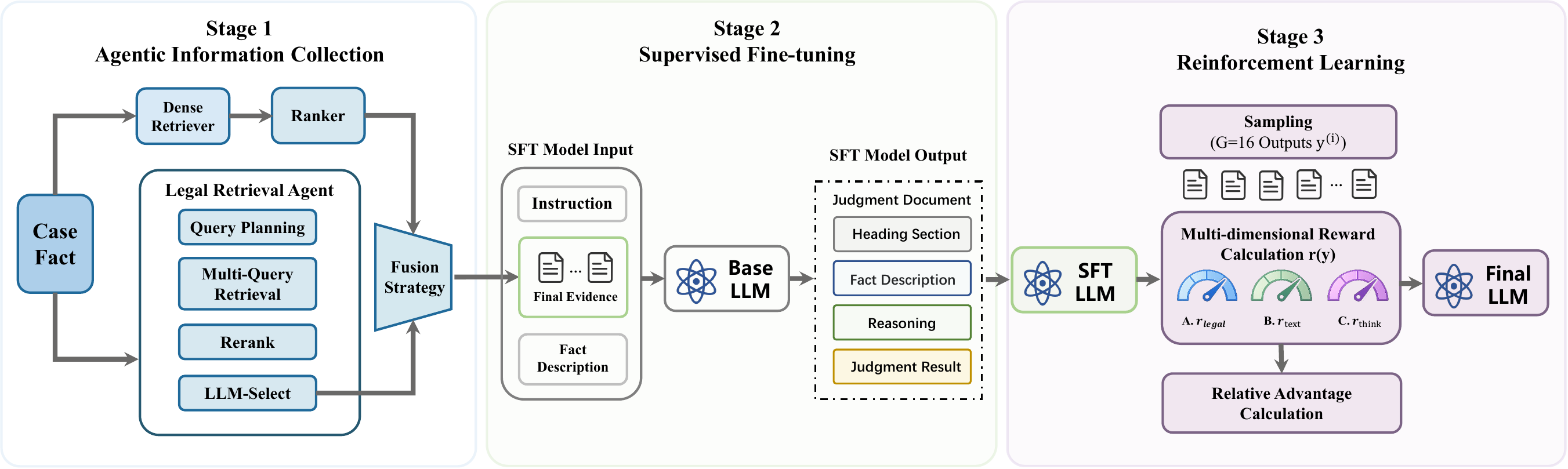}
\caption{An illustration of our proposed framework.}
\label{fig:method}
\end{figure*}

\section{Problem Formulation}
We formalize judgment document generation as a two-stage task. Given a fact description $f$ and a legal corpus $\mathcal{D} = \{d_1, d_2, \ldots, d_N\}$ containing statutes and precedents, the goal is to generate a judgment document $j$ that is legally sound and logically coherent. The process consists of:

\begin{itemize}[leftmargin=*]
\item \textbf{Legal Information Collection:} A retrieval system identifies a relevant evidence set $\mathcal{E}_f \subseteq \mathcal{D}$ containing the specific statutes and precedents pertinent to the facts $f$.
\item \textbf{Judgment Generation:} A generative system synthesizes the judgment $j$ conditioned on the fact description $f$ and the retrieved evidence $\mathcal{E}_f$, ensuring the output strictly adheres to judicial reasoning and sentencing standards.
\end{itemize}

\section{Methodology}
\label{sec:method}

We propose \textbf{Judge-R1}, a framework designed to synthesize legally sound judgment documents $j$ grounded in retrieved evidence $\mathcal{E}_f$. To ensure the generated judgments are both accurate in citing statutes and compliant with judicial reasoning, Judge-R1 operates through a three-stage pipeline (shown in Figure~\ref{fig:method}):
\begin{enumerate}[leftmargin=*]
    \item \textbf{Hybrid Information Collection:} A dual-route retrieval module that combines a standard dense backbone with an LLM-driven agent to construct a high-precision evidence set $\mathcal{E}_f$ from the corpus $\mathcal{D}$.
    \item \textbf{Supervised Fine-Tuning (SFT):} The generator $G_\theta$ takes the instruction, facts $f$, and retrieved laws $\mathcal{E}_f$ as input to generate the full judgment document $j$, establishing the foundational structure and formatting conventions.
    \item \textbf{Group Relative Policy Optimization (GRPO):} We further align the model using reinforcement learning with a multi-dimensional legal rubric, rewarding correctness and logic over mere surface-level mimicry.
\end{enumerate}

\subsection{Information Collection}\label{sec:method:collection}
\noindent\textbf{Overview.} We aim to retrieve a concise and high-recall set of case-relevant legal evidence $\mathcal{E}_f$, including applicable statutes and precedents, from the fact description $f$. 
For the standard route, we treat the fact description $f$ as the initial query $q$.
To balance coverage with precision, we employ a hybrid system fusing a robust neural backbone with a GRPO-enhanced reasoning agent.

\paragraph{Neural Backbone with Hard-Negative Mining.} 
The foundation is a two-stage retrieval pipeline. 
A bi-encoder encodes $q$ and $d$ into embeddings for dot-product retrieval. Crucially, we enhance discrimination using contrastive learning with {static hard negatives} mined from top-ranked near neighbors (excluding positives). 
Top candidates are reranked by a BERT-based cross-encoder. To align training with inference difficulty and prevent leakage, we construct training triples using a {K-fold mining strategy}, where negatives for one fold are retrieved by a model trained on disjoint folds.

\paragraph{GRPO-Enhanced Retrieval Agent.} 
Standard retrieval struggles with verbose, multi-issue narratives. We introduce an agentic workflow optimized via {Group Relative Policy Optimization (GRPO)~\cite{guo2025deepseek}} to align planning with downstream utility:

\begin{enumerate}[leftmargin=*,itemsep=1pt,topsep=1pt,parsep=0pt]
\item \textbf{QueryGen (Planning):} 
The agent decomposes $f$ into $m$ sub-queries targeting heterogeneous legal aspects, such as constitutive elements and sentencing factors. 
We apply GRPO here to reward retrieval effectiveness rather than merely linguistic quality. 
Specifically, each generated query is executed by the retriever, and the resulting ranking is evaluated against the gold legal evidence labels provided by the JuDGE dataset using MRR and Recall@50.

\item \textbf{Multi-View Recall \& Rerank:} 
We aggregate the top-$k$ results retrieved for all generated sub-queries and rerank them with the cross-encoder, forming a high-recall candidate pool.

\item \textbf{LLM-Select (Filtering):} 
Given the aggregated candidate pool, the agent selects a compact subset of applicable statutes and precedents. 
The selector is restricted to retrieved candidates. 
We train this stage with GRPO using rewards based on the gold evidence labels in JuDGE, balancing early precision ($P@5$), recall over the gold evidence available in the candidate pool, and a soft penalty on the number of selected statutes. 
The length penalty is tuned on the validation set to prevent both over-pruning and excessive selection. 
When the selected set contains fewer than $N_{\min}$ statutes, we supplement it with the highest-ranked remaining candidates from the reranker.
\end{enumerate}

\paragraph{Multi-route Fusion.} 
We combine two complementary retrieval routes via weighted Reciprocal Rank Fusion (as shown in Figure~\ref{fig:method}). 
The standard route directly retrieves candidates from the original case facts and reranks them with the cross-encoder. 
The agentic route first plans multiple legal queries, retrieves and reranks candidates for these queries, and then applies LLM-based selection. 
We set $w_{\text{agent}}{=}2.0$ and $w_{\text{std}}{=}1.0$ based on validation performance, prioritizing agent-selected evidence while preserving the robustness of the standard route.

\subsection{Judgment Document Generation}\label{sec:method:generation}
\noindent\textbf{Overview.} The generation module synthesizes a comprehensive judgment document $j$ conditioned on the fact description $f$ and the retrieved evidence set $\mathcal{E}_f$. The objective is to produce a document that not only reaches the correct verdict but also strictly adheres to judicial reasoning standards. To achieve this, we employ a two-stage optimization strategy: Supervised Fine-Tuning (SFT) for structural format alignment, followed by Group Relative Policy Optimization (GRPO) to enforce legal correctness and reasoning fidelity.

\paragraph{Supervised Fine-Tuning (SFT)}
We first adapt the LLM to the judicial domain using retrieval-augmented training instances. The input sequence is standardized as a concatenation of the instruction, retrieved evidence $\mathcal{E}_f$, and facts $f$, while the target output is the complete judgment document $j$. 
We optimize the model with the standard next-token prediction objective, minimizing the negative log-likelihood over the target judgment tokens.
This stage primarily focuses on stabilizing the generation format, ensuring the model learns to organize the judgment $j$ into distinct sections (e.g., fact acknowledgment, reasoning, and sentencing) and to explicitly cite the provided statutes. We use LoRA for parameter-efficient tuning, enabling the model to internalize the specific writing style of legal judgments while preserving its general reasoning capabilities. This warm-up phase provides a robust policy initialization for subsequent reinforcement learning.

\paragraph{Group Relative Policy Optimization (GRPO)}
While SFT establishes the document structure, it does not inherently guarantee the correctness of the legal conclusion. To align the generator with high-stakes judicial criteria, we further optimize it using GRPO. Unlike PPO-based RL fine-tuning, which typically relies on an additional value model, GRPO computes a group-relative baseline from the rewards of sampled outputs, reducing the overhead of training an extra critic model. This formulation is well suited to judgment document generation, where prior studies have established explicit evaluation dimensions such as legal correctness, structural professionalism, and logical consistency. These dimensions can be naturally operationalized as rubric-based rewards for policy optimization.
For each input, we sample a group of $G$ candidate judgments $\{j^{(i)}\}_{i=1}^G$ from the rollout policy and use them to update the generator policy $\pi_\theta$. Each candidate is evaluated by the reward rubric described below. 
The group-relative advantage used for policy optimization is computed as:
\begin{equation}
    A^{(i)} = \frac{r(j^{(i)}) - \mu_G}{\sigma_G + \epsilon}, 
    \quad 
    \mu_G = \frac{1}{G}\sum_{k=1}^G r(j^{(k)}).
\end{equation}
Here, $r(\cdot)$ denotes the scalar reward assigned by the rubric, $\mu_G$ and $\sigma_G$ denote the mean and standard deviation of the rewards within the sampled group, respectively, and $\epsilon$ is a small constant for numerical stability.

\paragraph{Rubric-Aware Reward Design.}
Since GRPO directly optimizes the generator according to scalar rewards, the reward function should reflect the domain-specific criteria of judgment quality.
We therefore design a rubric-aware reward that evaluates legal correctness, structural professionalism, and reasoning quality. 
The total reward $r(j)$ is defined as a weighted sum of three components:
\begin{equation}
    r(j) = w_1 \cdot r_{{legal}}(j) + w_2 \cdot r_{{struct}}(j) + w_3 \cdot r_{{logic}}(j),
    \label{eq:reward}
\end{equation}
where the weights $w_1$, $w_2$, and $w_3$, together with internal sub-weights, are configurable hyperparameters (see Sec.~\ref{sec-setup}).

\begin{itemize}[leftmargin=*]
    \item \textbf{Legal Correctness ($r_{{legal}}$):} This component evaluates whether the judgment reaches a legally correct conclusion by comparing extracted legal entities, including statutes, charges, sentencing terms, and fines, against the ground truth. Specifically, we calculate the $F_1$ score over the extracted statutory citations and charges. 
    Following the setting of JuDGE~\cite{su2025judge}, for numerical sentencing terms and fines, we employ the matching score
    \[
    S(A,B) = \max\left(0, 1 - \frac{|A-B|}{\max(A,B,1)}\right),
    \]
    where $A$ and $B$ denote the predicted and ground-truth values, respectively. We assign a zero score for categorical conflicts, such as acquittals versus convictions.
   
    \item \textbf{Structural Professionalism ($r_{{struct}}$):} Beyond legal correctness, a judgment document should also follow canonical court formats. We segment the document into \textit{reasoning} and \textit{judgment} sections and compute the average BERTScore between the generated segments and their corresponding ground-truth references. This encourages the generated document to align with reference judgments in both section-level content and professional expression. If a section is missing or cannot be extracted, we assign a score of $0$ to that section.

    \item \textbf{Reasoning Quality ($r_{{logic}}$):} 
    To discourage shallow or shortcut reasoning patterns, we evaluate the intermediate reasoning trace within the \texttt{<think>} tags. 
    Specifically, we define $r_{\text{logic}} = \min(1, S_{\text{len}} + S_{\text{rep}})$, where $S_{\text{len}}$ rewards reasoning traces with an appropriate length, and $S_{\text{rep}}$ gives higher scores to traces with less structural repetition.
    All interval thresholds and reward increments are treated as configurable hyperparameters (see Sec.~\ref{sec-setup}) to discourage degenerate behaviors such as circular reasoning, excessive repetition, or overly short reasoning traces.    
\end{itemize}

\begin{table}
\centering
\caption{Retrieval performance for information collection. Metrics include Precision (P), and Recall (R) at various cutoffs (5 and 10). The best results are in bold.}
\label{tab:retrieve}
\begin{tabular}{ccccc}
\toprule
                       & \textbf{P@5}    & \textbf{P@10}   & \textbf{R@5}    & \textbf{R@10}    \\
                       \midrule
\textbf{TF-IDF}        & 0.1625          & 0.1200          & 0.2096          & 0.3074           \\
\textbf{BM25}          & 0.1501          & 0.1024          & 0.1958          & 0.2649           \\
\textbf{Law Retriever} & {0.4535} & {0.3509} & {0.5529} & {0.8262} \\
\textbf{Agentic (Ours)} & \textbf{0.6383} & \textbf{0.3878} & \textbf{0.7806} & \textbf{0.9159} \\
\toprule
\end{tabular}
\end{table}

\begin{table*}
\centering
\setlength\tabcolsep{3pt} 
\caption{Main experimental results. We report accuracy for Penalty (Prison and Fine terms), and Recall, Precision, and F1 scores for Convicting and Referencing tasks. For the Reasoning and Judgment sections, we use METEOR (MET.) and BERTScore (BERTS.) as metrics. The best results for each model are highlighted in bold.}
\label{tab:main_results}
\begin{tabular}{cccccccccccccc}
\toprule
 & & \multicolumn{2}{c}{\textbf{Penalty Acc.}} & \multicolumn{3}{c}{\textbf{Convicting Acc.}} & \multicolumn{3}{c}{\textbf{Referencing Acc.}} & \multicolumn{2}{c}{\textbf{Reasoning Section}} & \multicolumn{2}{c}{\textbf{Judgment Section}} \\
\cmidrule(lr){3-4}\cmidrule(lr){5-7}\cmidrule(lr){8-10}\cmidrule(lr){11-12}\cmidrule(lr){13-14}
\textbf{Model} & \textbf{Method} & \textbf{Prison} & \textbf{Fine} & \textbf{Recall} & \textbf{Prec.} & \textbf{F1} & \textbf{Recall} & \textbf{Prec.} & \textbf{F1} & \textbf{MET.} & \textbf{BERTS.} & \textbf{MET.} & \textbf{BERTS.} \\
\toprule

\multirow{5}{*}{\shortstack[l]{\textbf{QW-2.5-3B}}} 
 & \textbf{Direct} & 0.6088 & 0.4639 & 0.8463 & 0.8426 & 0.8445 & 0.4666 & \textbf{0.7288} & 0.5689 & 0.4060 & 0.8059 & 0.3034 & 0.8030 \\
 & \textbf{ICL}    & 0.6332 & 0.4659 & 0.8792 & 0.8699 & 0.8730 & 0.4813 & 0.6785 & 0.5631 & 0.3890 & 0.7900 & 0.3895 & 0.8274 \\
 & \textbf{SFT}    & 0.6440 & 0.4971 & 0.8962 & 0.8985 & 0.8974 & 0.6130 & 0.7287 & 0.6659 & 0.5087 & 0.8696 & 0.5843 & 0.8872 \\
 & \textbf{RAG+SFT}   & 0.5825 & \textbf{0.4977} & 0.9102 & 0.9102 & 0.9102 & 0.5563 & 0.6167 & 0.5850 & 0.3896 & 0.7691 & 0.3776 & 0.8134 \\
 & \textbf{Judge-R1} & \textbf{0.6625} & 0.4857 & \textbf{0.9281} & \textbf{0.9278} & \textbf{0.9280} & \textbf{0.7589} & 0.6905 & \textbf{0.7231} & \textbf{0.5660} & \textbf{0.8853} & \textbf{0.6558} & \textbf{0.9036} \\
\midrule

\multirow{5}{*}{\shortstack[l]{\textbf{QW-3-4B}}} 
 & \textbf{Direct} & 0.5409 & 0.3454 & 0.9142 & 0.9125 & 0.9133 & 0.4978 & 0.6668 & 0.5700 & 0.4508 & 0.8298 & 0.3976 & 0.8215 \\
 & \textbf{ICL}    & 0.5909 & 0.3973 & 0.9321 & 0.9391 & 0.9256 & 0.4981 & 0.7199 & 0.5888 & 0.3873 & 0.7622 & 0.4350 & 0.8627 \\
 & \textbf{SFT}    & 0.6224 & 0.4443 & 0.9082 & 0.9079 & 0.9080 & 0.5753 & \textbf{0.7216} & 0.6402 & 0.5051 & 0.8741 & 0.5455 & 0.8829 \\
 & \textbf{RAG+SFT}   & 0.5732 & 0.4648 & 0.9321 & 0.9358 & 0.9340 & 0.7062 & 0.6470 & 0.6753 & 0.4460 & 0.8098 & 0.5368 & 0.8719 \\
 & \textbf{Judge-R1} & \textbf{0.6573} & \textbf{0.5144} & \textbf{0.9391} & \textbf{0.9408} & \textbf{0.9400} & \textbf{0.7739} & 0.7052 & \textbf{0.7379} & \textbf{0.5363} & \textbf{0.8855} & \textbf{0.6476} & \textbf{0.9042} \\
\bottomrule
\end{tabular}
\end{table*}

\section{Experimental Setup}\label{sec-setup}
\noindent \textbf{Benchmark} We conduct our experiments on the \textbf{JuDGE} benchmark~\cite{su2025judge}, a dataset specifically constructed to evaluate the generation of complete judgment documents based on factual descriptions. The dataset comprises 2,505 fact-judgment pairs, officially partitioned into 2,004 training instances and 501 test instances. 

\noindent \textbf{Baselines} For the baselines of the {Information Collection} stage, we employ traditional sparse retrievers including {TF-IDF} and {BM25}. We also include the {Law Retriever}~\cite{su2025judge}, the previous state-of-the-art dense bi-encoder, which is explicitly trained via contrastive learning to align case facts with statutes. For the {Judgment Generation} stage, we adopt three baselines from the JuDGE benchmark~\cite{su2025judge}: (1) {Few-shot ICL}, which prompts LLMs with fact-judgment exemplars; (2) {SFT}, which directly optimizes the model on ground-truth documents; and (3) {RAG+SFT}, specifically the Multi-Source RAG (MRAG) framework that augments SFT with external knowledge from both Law and Case Retrievers.

\noindent \textbf{Implementation Details} We utilize Qwen2.5-3B-Instruct and Qwen3-4B-Thinking as our backbone models. The dense retriever is initialized with Chinese-RoBERTa-WWM and optimized using AdamW (LR=$2\text{e-}5$, fp16), trained on the training set from the JuDGE benchmark, while LLM fine-tuning employs a learning rate of $2\text{e-}5$ (bf16). For the GRPO stage, we train for 1 epoch (LR=$5\text{e-}6$, KL $\beta=0.05$) with $16$ generations per input, using a reward function that aggregates $r_{\text{legal}}$, $r_{\text{struct}}$, and $r_{\text{reason}}$ with weights of $0.6$, $0.3$, and $0.1$, respectively. All experiments are conducted on 4 NVIDIA A800 (80GB) GPUs using default Hugging Face configurations.

\section{Experimental Results}\label{sec:exp_results}

\subsection{Analysis of Legal Information Collection}

The evaluation results on the Information Collection stage demonstrate the decisive advantage of our proposed Judge-R1 framework over traditional retrieval baselines. As indicated by the performance comparison, the Agentic Legal Information Collection mechanism substantially improves the retrieval of relevant legal evidence, including statutes and precedents. Two key trends emerge from the analysis. First, Judge-R1 exhibits significantly higher recall compared to both lexical matching methods and the previous SOTA dense retrieval method. This indicates that the agent's ability to dynamically decompose complex case facts into precise queries effectively captures obscure legal points that static semantic matching often overlooks. Second, the multi-stage filtering mechanism ensures high precision by effectively suppressing noise in the retrieval space and distinguishing between legally distinct but textually similar statutes. Overall, these observations validate that moving beyond passive retrieval to an agentic, planning-based approach is essential for handling complex legal reasoning tasks.

\subsection{Analysis of Judgment Document Generation}

As shown in Table ~\ref{tab:main_results}, Judge-R1 achieves the best overall performance across both model backbones, obtaining the strongest results on most key metrics, especially citation recall/F1 and document-level generation quality.
First, regarding {legal decision-making} (Penalty and Convicting), Judge-R1 achieves substantial gains over the strong SFT baseline. This indicates that while SFT captures surface-level stylistic patterns, our RL alignment via GRPO effectively reinforces logical rigor, enabling the model to deduce more accurate sentencing terms and conviction results.
Second, with respect to {citation and evidence utilization} (Referencing), our method outperforms standard RAG approaches. The superior Precision and F1 scores demonstrate that the proposed Agentic Legal Information Collection mechanism reduces retrieval noise and aligns retrieved evidence more precisely with case facts than varying semantic matching.
Finally, for {document generation quality} (Reasoning and Judgment sections), Judge-R1 secures the highest scores in both METEOR and BERTScore. This confirms that our framework not only improves legal accuracy but also enhances the structural integrity and coherence of the generated text, successfully synthesizing complex legal logic into professional-grade judgments.

\section{Related Work}

Early research in legal intelligence primarily focused on {Legal Judgment Prediction (LJP)}, which formulates judicial reasoning as a classification problem, such as predicting charges, applicable articles, or sentencing terms from case facts~\cite{luo2017learning,hu2018few,li2019mann,kang2019creating,chen2019charge,nigam2024rethinkinglegaljudgementprediction}. 
Although these methods are effective for outcome prediction, they do not aim to generate complete judicial documents with explicit legal reasoning and formal writing structures. 
More recently, the {JuDGE} benchmark~\cite{su2025judge} introduced {Judgment Document Generation} as a more comprehensive task, where LLMs are required to produce full judgment documents conditioned on case facts and legal evidence. 
This task goes beyond predicting legal outcomes, as it requires the model to identify relevant legal authorities, organize procedural and factual descriptions, and produce reasoning that supports the final judgment results.

From a broader perspective, judgment document generation is closely related to Retrieval-Augmented Generation (RAG), which enhances LLMs by retrieving task-relevant information from external sources and using it to ground model generation~\cite{lewis2020retrieval,dong2025decoupling,tu2025robust,su2025dynamic,tu2026analytical,su2025sigirap}. 
Prior studies have shown that RAG is useful for mitigating hallucinations~\cite{wang2026joint,su2025towards,su2024unsupervised}, supporting timely knowledge updates~\cite{wang2025decoupling,wang2025knowledge,wang2025decoupling}, and adapting LLMs to specialized domains without expensive full-parameter retraining~\cite{su2025judge,su2024stard,wang2024lekube}. 
A typical RAG pipeline first retrieves relevant documents from a large-scale corpus~\cite{robertson2009probabilistic,su2024wikiformer,fang2024scaling,su2025pre,tu2026generalized,su2025surge}, and then conditions the generator on the retrieved evidence to produce a grounded response. 
Recent work has further extended this paradigm in several directions, including dynamic RAG~\cite{jiang2023active,su2024dragin,su2024mitigating}, graph RAG~\cite{edge2024local}, parametric RAG~\cite{su2025parametric,tan2025dynamic}, and agentic RAG~\cite{jin2025search,su2026skill}.

Despite these advances, judgment document generation imposes stricter requirements than standard RAG-based question answering or summarization. 
In legal scenarios, retrieved materials are not merely supporting context but also serve as legal authorities that must be selected, cited, and integrated into a valid chain of judicial reasoning. 
Existing baselines for judgment document generation largely rely on standard RAG and Supervised Fine-Tuning (SFT), but they often struggle to retrieve precise statutes and precedents under legal ambiguity~\cite{goebel2023summary,kim2022coliee,rabelo2022overview,su2024stard}, and lack objective-driven feedback to penalize legally unsupported reasoning or hallucinated citations. 
To address these limitations, Judge-R1 combines agentic planning for legal information collection with rubric-driven reinforcement learning, enabling the model to improve both evidence grounding and consistency in legal reasoning.

\section{Conclusion}

In this paper, we propose \textbf{Judge-R1}, a framework that integrates agentic legal information collection with rubric-driven reinforcement learning for judgment document generation. 
By improving both legal evidence collection and reasoning-grounded generation, Judge-R1 addresses key bottlenecks in judgment document generation and achieves state-of-the-art performance on the JuDGE benchmark. 
In future work, we plan to extend this paradigm to broader legal reasoning tasks and investigate more efficient alignment strategies for legal-domain LLMs.

\section{Acknowledgement}
This work is supported by the Research Project of Quan Cheng Laboratory, China (Grant No. QCL20250105).


\bibliographystyle{ACM-Reference-Format}
\balance
\bibliography{sample-base}

\end{document}